\documentclass[pdflatex,sn-mathphys-num]{sn-jnl}% Math and Physical Sciences Numbered Reference Style 
\usepackage{graphicx}%
\usepackage{multirow}%
\usepackage{amsmath,amssymb,amsfonts}%
\usepackage{amsthm}%
\usepackage{mathrsfs}%
\usepackage[title]{appendix}%
\usepackage{xcolor}%
\usepackage{textcomp}%
\usepackage{manyfoot}%
\usepackage{booktabs}%
\usepackage{algorithm}%
\usepackage{algorithmicx}%
\usepackage{algpseudocode}%
\usepackage{listings}%
\usepackage{makecell}

\theoremstyle{thmstyleone}%
\theoremstyle{thmstyletwo}%

\theoremstyle{thmstylethree}%

\begin{document}

\title[Article Title]{Leveraging Large Language Models for Fuzzy String Matching in Political Science}

\author{\fnm{Yu} \sur{Wang}}\email{yuwang.aiml@gmail.com}


\affil{\orgdiv{Institute for Advanced Study in Social Sciences}, \orgname{Fudan University}, \orgaddress{\city{Shanghai}, \postcode{200433}, \country{China}}}


% \affil[3]{\orgdiv{Department}, \orgname{Organization}, \orgaddress{\street{Street}, \city{City}, \postcode{610101}, \state{State}, \country{Country}}}


\abstract{Fuzzy string matching remains a key issue when political scientists combine data from different sources. Existing matching methods invariably rely on string distances, such as Levenshtein distance and cosine similarity. As such, they are inherently incapable of matching strings that refer to the same entity with different names such as ``JP Morgan'' and ``Chase Bank'', ``DPRK'' and ``North Korea'', ``Chuck Fleischmann (R)'' and ``Charles Fleischmann (R)''. In this letter, we propose to use large language models to entirely sidestep this problem in an easy and intuitive manner. Extensive experiments show that our proposed methods can improve the state of the art by as much as 39\% in terms of average precision while being substantially easier and more intuitive to use by political scientists. Moreover, our results are robust against various temperatures. We further note that enhanced prompting can lead to additional performance improvements.}

\keywords{Large language models, ChatGPT, prompt engineering, text as data}



\maketitle

% unsupervised_semi_supervised_visual_frames,face_detection_political_figures_news_archives,video

% finetune_pa,cross_domain
\section*{Introduction}
Political scientists frequently deal with many variables in their research and, as part of the data preparation process, need to merge datasets from different sources using a shared identifier. For example, North Korea is represented as ``Korea, Dem. People's Rep.'' in economic data from the World Bank, as ``Korea, Dem. People's Rep. of'' in trade data from the International Monetary Fund, and as ``Korea North'' in polity scores from the Center for Systemic Peace~\citep{swy}. This is sometimes referred to as the fuzzy string matching problem. Existing solutions to the fuzzy string matching problem invariably operate at the string level. Such solutions include edit distance, longest common substring, and bag-of-words similarities. Edit distance is good at matching cases with minor typos, longest common substring is designed to be sensitive to even the smallest letter differences, and bag-of-words similarities are robust against word orders. More recently, researchers have proposed a method that utilizes all these different distances as separate features to build a random forest. The new method further involves human annotation in the loop (HITL) to overcome predictive deficiencies~\citep{fuzzy}. Their method shows improvement over using the single solutions above. Nevertheless, given that all these individual features are ultimately based on matching raw strings, the resulting random forest still cannot overcome the inherent difficulty of matching entities expressed in totally different terms. For example, the model performs poorly in such cases as ``DPRK'' and ``North Korea'', ``JP Morgan'' and ``Chase Bank'',  ``Chuck Fleischmann (R)'' and ``Charles Fleischmann (R)'' At the same time, the involvement of human annotation in the loop increases the cost and slows down the velocity of model development.

In this paper we propose a semantics-based solution that leverages large language models to overcome the limitations noted above. Specifically, we use ChatGPT as a zero-shot classifier for fuzzy string matching. ChatGPT is known to contain world knowledge~\citep{aug,kola,medical_science}, so we expect them to see the semantic similarity between ``DPRK'' and ``North Korea'' and between ``JP Morgan'' and ``Chase Bank.'' As a sanity check, we did ask ChatGPT the following question: ``Does DPRK and North Korea refer to the same country? Yes or no?'' and got the correct answer. At the same time, ChatGPT uses byte-pair encoding (BPE), a subword encoding mechanism~\citep{bpe}, and consequently is relatively robust against typos in individual words. Indeed, researchers have already identified several areas where ChatGPT excels in zero-shot learning~\citep{zero-shot,zero-shot-emnlp2022}, which motivates us to experiment with using ChatGPT as a zero-shot classifier for fuzzy string matching. Extensive experiments offer strong evidence that the zero-shot ChatGPT outperforms the best performing string matching methods by a substantial margin, while being substantially easier and more intuitive to use.  Our results are robust against the use of different temperatures. In addition, we observe that improved prompting could lead to performance gains.
\section*{Results}

\begin{figure*}[!t]
\centering
\includegraphics[width=380px]{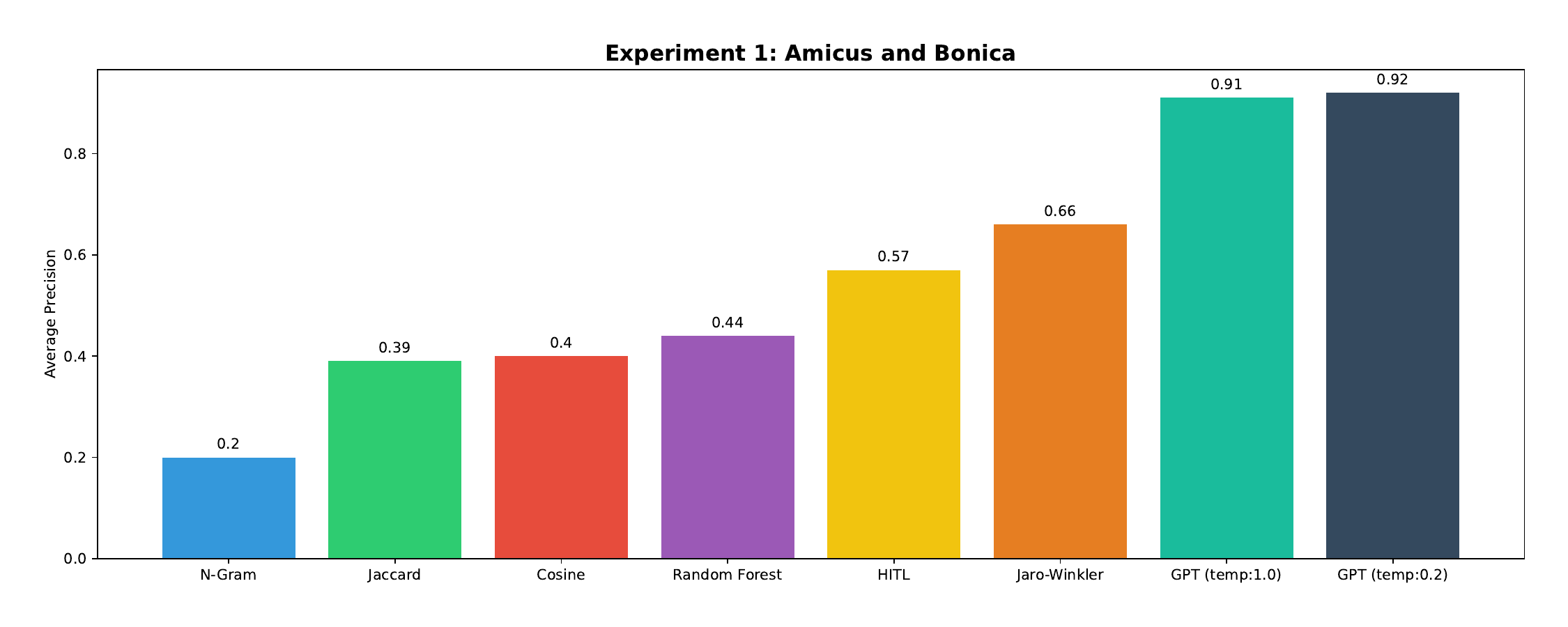}
\caption{ChatGPT substantially outperforms character-based matching methods. The zero-shot ChatGPTs, at temperatures 0.2 and 1, both outperform character-based methods by a large margin.}
\label{fig1}
\end{figure*}

We use two main datasets (n = 4,500) that cover both organizations and politicians in the United States from~\citep{fuzzy}. The first dataset, \textit{Amicus and Bonica}, focuses on matching organizations and has 4,000 samples. The second dataset, Incumbent Voting, has 500 samples and centers around matching political candidates. The second dataset is built from a larger set of mismatched named entities. That set is first sorted using a random forest-based model from~\citep{fuzzy} and the top five hundred most likely matches come to form the \textit{Incumbent Voting} dataset.

%We split the first dataset into train, validation and test sets the same as is done in~\cite{fuzzy} to train RoBERTa-large. 

%The second dataset, \textit{Incumbent Voting}, is used exclusively for inference with string-based methods, of which Base Model is trained on the first dataset, RoBERTa finetuned on the first dataset, and the zero-shot ChatGPT.

%For each subplot, the y-axis measures precision and the x-axis measures recall. In the top row, 

In Figure~\ref{fig1}, we report our result on the \textit{Amicus and Bonica} dataset in terms of average precision (AP) scores. we plot the results for N-Gram Overlap, Jaccard, Cosine, Random Forest, HITL, Jaro-Winkler, ChatGPT-T10 and ChatGPT-T02. HITL refers to the human-in-the-loop algorithm proposed by~\cite{fuzzy} and is built on top of a random forest. HITL differs from the random forest model in that it involves actual human annotation in the loop. The first fix bars use the results from~\cite{fuzzy}. ChatGPT-T10 refers to ChatGPT with a temperature of 1.0. ChatGPT-T02 refers to ChatGPT with a temperature of 0.2.

We observe that HITL~\citep{fuzzy} and Jaro-Winkler perform the best among the existing models, with an average precision score of 0.57 and 0.66, respectively. n-Gram Overlap performs the worst with an average precision of 0.2. Note that both Jaccard similarity and cosine similarity are composition-based measures. Jaccard similarity measures the ratio of in-common characters to total characters. Cosine similarity measures the cosine of the angle between two letter frequency vectors. N-Gram Overlap refers to the longest common substring. Jaro-Winkler is one type of edit distance is calculated using the R package RecordLinkage~\citep{fuzzy}. In comparison, the zero-shot ChatGPT attains an average precision score of 0.91 at a temperature setting of 1.0 and improves slightly to a score of 0.92 when the temperature is reduced to 0.2. This represents a gain of 39\% for ChatGPT over existing methods. This is particularly noteworthy given that, unlike HITL, ChatGPT does not use any training samples in the current setting.

\begin{figure*}[!t]
\centering
\includegraphics[width=380px]{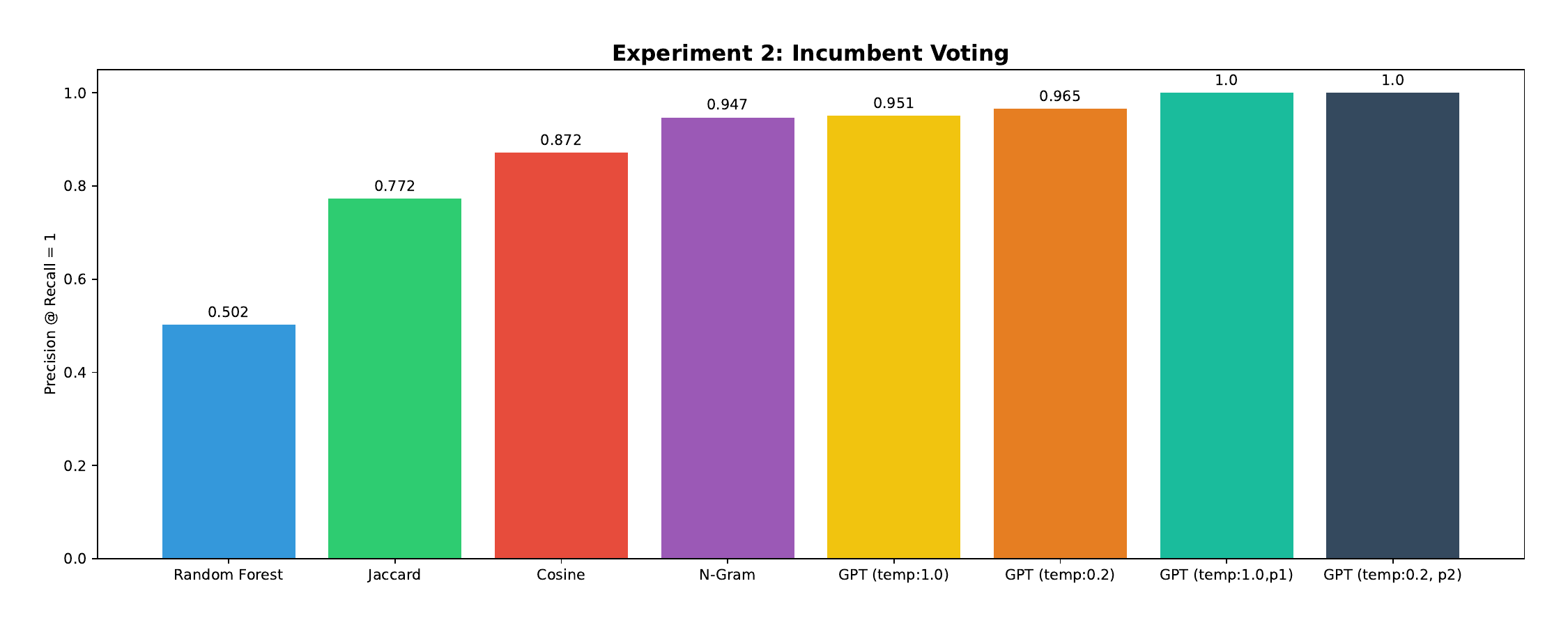}
\caption{ChatGPT again outperforms character-based matching methods. In particular, when we provide more context in the prompt (p2), zero-shot ChatGPTs, at temperatures 0.2 and 1.0, both outperform character-based methods by a large margin and achieve 100\% precision.}
\label{fig2}
\end{figure*}

Figure~\ref{fig2} shows results for the Incumbent Voting dataset in terms of precision at recall = 1.0, where our goal is to identify all the matching politicians (100\% recall) in the dataset with the highest precision. As our baselines, we have results for Jaccard, Random Forest (trained on the \textit{Amicus and Bonica} dataset), and Cosine. The random forest model is the same model as used in Figure~\ref{fig1}. Here it is directly applied to \textit{Incumbent Voting}, which is considered out of domain~\citep{cross_domain}. Besides the baseline models, we report results from zero-shot ChatGPT, with two different temperatures and different prompting schemes.

We observe that n-gram performs best among the character-based methods, achieving a precision of 0.947. By comparison, ChatGPT achieves a precision of 0.951 with temperature at 1.0 and 0.966 with temperature at 0.2, both using plain prompting. When we add additional information about the entity pairs, including what `(R)' and `(D)' stand for as well as the fact that all those entities are Congressional candidates, we observe that precision can reach 1.0 at both temperatures, which represents another 6\% gain over the best performing character-based method, n-gram.

\section*{Discussion}
\subsubsection*{Performance} Fuzzy string matching is a long-standing problem in political science. So far, existing methods have exclusively focused on character-level calculations. In this letter, we have proposed a semantics-based matching methods that leverage the latest large language models. We have shown that zero-shot ChatGPT outperforms the best existing methods by as much as 39\% in the \textit{Amicus and Bonica} dataset. In the \textit{Incumbent Voting}, where records represent the most probable matching pairs as ranked by a base random forest model, ChatGPT was able to 
reach a precision of 1.0 with detailed prompting. 

\subsubsection*{Ease of use} In addition to its exceptional performance, we note that another advantage of using zero-shot ChatGPT is its ease of use. Among the existing character-based matching methods, some of them, such as random forest, HITL, and n-gram, can get quite complicated. By contrast, ChatGPT and other similar LLMs only require a simple and intuitive command in a natural language, which drastically lowers the technical barriers.
  
\subsubsection*{Running Time} In terms of running time, we note that running zero-shot inference using ChatGPT-4 for 4,000 samples takes about 3,545 seconds (max among our multiple runs). While this is doable for small datasets of similar sizes, running time for larger datasets could become a bottleneck. Potential solutions to speeding up the process include (1) parallelizing the API calls, (2) making calls in batches, and (3) using the ChatGPT-3, which is faster, instead of ChatGPT-4. Alternatively, researchers could adopt a two-stage approach, where at the first stage a fast but less accurate model, e.g. Cosine Similarity, is used to sort the entity pairs in terms of matching probability and at the second stage ChatGPT is applied to grading only the top ranked entity pairs. This is similar to the experiment on \textit{Incumbent Voting} in this study.

\section*{Materials and methods}
\subsection*{Data}
We use the main datasets from~\cite{fuzzy}: \textit{Amicus and
Bonic} and \textit{Incumbent Voting}. In \textit{Amicus and
Bonic}, strings representing campaign donors~\citep{bonica} are matched to strings that represent amicus curiae cosigners~\citep{amicus}. The training set contains 71,043 string pairs in total. Please refer to Table~\ref{1} for an illustration of the training data and see the replication package for more details. In terms of the distribution of positive labels and negative labels, we note that only 357 string pairs refer to the same entity. Given that the positive samples represent only 0.5\% of the training set, the dataset is severely imbalanced. For related studies on rare event classification (e.g., civil wars), readers can refer to~\cite{predicted_acc1} and ~\cite{predicted_acc2}. The test set contains 4,000 string pairs. In our experiments, we use the test set exclusively.

In the \textit{Incumbent Voting} dataset, strings representing names of the Congressional candidates hand-typed by respondents are matched to strings that represent the standard version of candidates' names. The data points in~\textit{Incumbent Voting} are part of the 5,356 mismatched pairs in total.~\cite{fuzzy} build~\textit{Incumbent Voting} by first ranking all these mismatched pairs by the random forest model trained on \textit{Amicus and
Bonic}, select the top 500 most likely matches, and then label them.

\begin{table}[h]
\caption{Sample data from Datasets \textit{Amicus and Bonica} and \textit{Incumbent Voting}. Capitalization is kept the same as in the original dataset.}
\label{1}
\begin{tabular}{lll}
\hline
Left                            & Right                                     & Label \\
\hline\hline
JPMORGAN CHASE BANK NA            & jp morgan chase  & 1    \\   
roadway express inc & motion picture assn & 0 \\
Chuck Fleischmann (R) & Charles Fleischmann (R) & 1\\
Brad Sherman (D) & Howard Berman (D) & 0  
\\\hline
\end{tabular}
\end{table}

% \subsubsection*{Discriminative RoBERTa}
% For the RoBERTa model, we finetune the large version with 340 million parameters for binary classification~\citep{roberta,finetune_pa}. During training and inference, we concatenate each string pair into a single string input with the separator token, $<$s$>$,  as follows:

% %"amicus"] + " </s> "

% \[ amicus\:string +  `` </s> " +  bonica\:string\]

% Given the input sequences are relatively short, we set the maximum sequence length to 50 in an effort to speed up the training and inference process~\citep{pretrained_topic_classification}. We split the training set from~\citep{fuzzy} into a training set (63,938 or 90\% of the original training set) and a validation set (7,105 or 10\%), and use the test set with 4,000 samples as it is. We set the training batch size to 128 and the evaluation batch size to 512.  We apply a learning rate of 1e-5, finetune the RoBERTa-large model for 4 epochs, and use precision on the validation set as the criterion for selecting the optimal checkpoint.

\subsubsection*{Methods}
When calling the ChatGPT model, we do not use the training set. Instead, we build zero-shot prompts and rely on ChatGPT to make inferences on the test set directly~\citep{zero-shot}. Specifically, we use ChatGPT-4 version, gpt-4-0613, and use the OpenAI API for chat completion.\footnote{For an introduction to the API, please see https://platform.openai.com/docs/quickstart?context=python. GPT-4 was accessed online on Oct 28, 2023. For coding details, please refer to our replication package.} For our first experiment, we provide the prompt in a standardized format as follows, where entity\_a and entity\_b refer to the two entity strings, respectively:\\

\setlength{\leftskip}{1cm}
\setlength{\rightskip}{1cm}
\noindent\textit{How confident are you that the following entities, {entity\_a} and {entity\_b}, refer to the same entity, allowing for the possibility of minor typos?}

 \noindent\textit{Please return your confidence in the range of 0 and 1 only and no other words.}\\
 
\setlength{\leftskip}{0cm} 
\setlength{\rightskip}{0cm}

\noindent For our second experiment (Incumbent Voting), besides using the standard prompt, we also enrich the prompt with more information about each entity pair. For example, all the named entities are Congressional candidates. `R' stands for Republican in `Walter B. Jones (R)'.\\

\setlength{\leftskip}{1cm}
\setlength{\rightskip}{1cm}

\noindent\textit{You are a helpful and knowledgeable assistant. You will be given two entities. Both entities refer to Congressional candidates, where `R' stands for Republican and `D' stands for `Democrat'.}

\vspace{0.5cm}

\noindent\textit{How confident are you that the following entities, {entity\_a} and {entity\_b}, refer to the same entity, allowing for the possibility of minor typos?}

 \noindent\textit{Please return your confidence in the range of 0 and 1 only and no other words.}\\
 
\setlength{\leftskip}{0cm} 
\setlength{\rightskip}{0cm}

We make an individual call for each entity pair and then parse the returned response into a certainty score, which is a float number between 0 and 1. The higher the score, the more likely that the two entity strings refer to the same entity. We then use the ChatGPT score to sort and rank the likelihood of two strings referring to the same entity.

% Feb 21, 2024
% temperature 0.2
% end 1708546288.461317
% total 2268.943613052368
% 0.92

% Temperature: 0.2 
% Col chatGPT_score
% 0.9615384615384616

% \subsection*{Data, Materials, and Software Availability}
% Replication materials will be made available  at  the  Harvard  Dataverse, including data and the code.

% \section*{Competing interests}
% The author(s) declare no competing interests.

\section*{Data availability}
All data generated or analysed during this study are included in this published article and its supplementary information files.

\section*{Ethical approval}
This article does not contain any studies with human participants performed by any of the authors.

% \section*{Informed consent}
% This article does not contain any studies with human participants performed by any of the authors.

\bibliography{ir}% common bib file


\begin{thebibliography}{13}
\ifx \bisbn   \undefined \def \bisbn  #1{ISBN #1}\fi
\ifx \binits  \undefined \def \binits#1{#1}\fi
\ifx \bauthor  \undefined \def \bauthor#1{#1}\fi
\ifx \batitle  \undefined \def \batitle#1{#1}\fi
\ifx \bjtitle  \undefined \def \bjtitle#1{#1}\fi
\ifx \bvolume  \undefined \def \bvolume#1{\textbf{#1}}\fi
\ifx \byear  \undefined \def \byear#1{#1}\fi
\ifx \bissue  \undefined \def \bissue#1{#1}\fi
\ifx \bfpage  \undefined \def \bfpage#1{#1}\fi
\ifx \blpage  \undefined \def \blpage #1{#1}\fi
\ifx \burl  \undefined \def \burl#1{\textsf{#1}}\fi
\ifx \doiurl  \undefined \def \doiurl#1{\url{https://doi.org/#1}}\fi
\ifx \betal  \undefined \def \betal{\textit{et al.}}\fi
\ifx \binstitute  \undefined \def \binstitute#1{#1}\fi
\ifx \binstitutionaled  \undefined \def \binstitutionaled#1{#1}\fi
\ifx \bctitle  \undefined \def \bctitle#1{#1}\fi
\ifx \beditor  \undefined \def \beditor#1{#1}\fi
\ifx \bpublisher  \undefined \def \bpublisher#1{#1}\fi
\ifx \bbtitle  \undefined \def \bbtitle#1{#1}\fi
\ifx \bedition  \undefined \def \bedition#1{#1}\fi
\ifx \bseriesno  \undefined \def \bseriesno#1{#1}\fi
\ifx \blocation  \undefined \def \blocation#1{#1}\fi
\ifx \bsertitle  \undefined \def \bsertitle#1{#1}\fi
\ifx \bsnm \undefined \def \bsnm#1{#1}\fi
\ifx \bsuffix \undefined \def \bsuffix#1{#1}\fi
\ifx \bparticle \undefined \def \bparticle#1{#1}\fi
\ifx \barticle \undefined \def \barticle#1{#1}\fi
\bibcommenthead
\ifx \bconfdate \undefined \def \bconfdate #1{#1}\fi
\ifx \botherref \undefined \def \botherref #1{#1}\fi
\ifx \url \undefined \def \url#1{\textsf{#1}}\fi
\ifx \bchapter \undefined \def \bchapter#1{#1}\fi
\ifx \bbook \undefined \def \bbook#1{#1}\fi
\ifx \bcomment \undefined \def \bcomment#1{#1}\fi
\ifx \oauthor \undefined \def \oauthor#1{#1}\fi
\ifx \citeauthoryear \undefined \def \citeauthoryear#1{#1}\fi
\ifx \endbibitem  \undefined \def \endbibitem {}\fi
\ifx \bconflocation  \undefined \def \bconflocation#1{#1}\fi
\ifx \arxivurl  \undefined \def \arxivurl#1{\textsf{#1}}\fi
\csname PreBibitemsHook\endcsname

%%% 1
\bibitem[\protect\citeauthoryear{Stone et~al.}{2022}]{swy}
\begin{barticle}
\bauthor{\bsnm{Stone}, \binits{R.}},
\bauthor{\bsnm{Wang}, \binits{Y.}},
\bauthor{\bsnm{Yu}, \binits{S.}}:
\batitle{{{{Chinese Power and the State-Owned Enterprise}}}}.
\bjtitle{International Organization}
\bvolume{76}(\bissue{1}),
\bfpage{229}--\blpage{50}
(\byear{2022})
\end{barticle}
\endbibitem

%%% 2
\bibitem[\protect\citeauthoryear{Kaufman and Klevs}{2022}]{fuzzy}
\begin{botherref}
\oauthor{\bsnm{Kaufman}, \binits{A.R.}},
\oauthor{\bsnm{Klevs}, \binits{A.}}:
Adaptive fuzzy string matching: How to merge datasets with only one (messy) identifying field.
Political Analysis
(2022)
\end{botherref}
\endbibitem

%%% 3
\bibitem[\protect\citeauthoryear{Longpre et~al.}{2020}]{aug}
\begin{botherref}
\oauthor{\bsnm{Longpre}, \binits{S.}},
\oauthor{\bsnm{Wang}, \binits{Y.}},
\oauthor{\bsnm{DuBois}, \binits{C.}}:
{How Effective is Task-Agnostic Data Augmentation for Pretrained Transformers?}
Findings of the Association for Computational Linguistics: EMNLP 2020
(2020)
\end{botherref}
\endbibitem

%%% 4
\bibitem[\protect\citeauthoryear{Yu et~al.}{2024}]{kola}
\begin{botherref}
\oauthor{\bsnm{Yu}, \binits{J.}},
\oauthor{\bsnm{Wang}, \binits{X.}},
\oauthor{\bsnm{Tu}, \binits{S.}},
\oauthor{\bsnm{Cao}, \binits{S.}},
\oauthor{\bsnm{Zhang-Li}, \binits{D.}},
\oauthor{\bsnm{Lv}, \binits{X.}},
\oauthor{\bsnm{Peng}, \binits{H.}},
\oauthor{\bsnm{Yao}, \binits{Z.}},
\oauthor{\bsnm{Zhang}, \binits{X.}},
\oauthor{\bsnm{Li}, \binits{H.}},
\oauthor{\bsnm{Li}, \binits{C.}},
\oauthor{\bsnm{Zhang}, \binits{Z.}},
\oauthor{\bsnm{Bai}, \binits{Y.}},
\oauthor{\bsnm{Liu}, \binits{Y.}},
\oauthor{\bsnm{Xin}, \binits{A.}},
\oauthor{\bsnm{Lin}, \binits{N.}},
\oauthor{\bsnm{Yun}, \binits{K.}},
\oauthor{\bsnm{Gong}, \binits{L.}},
\oauthor{\bsnm{Chen}, \binits{J.}},
\oauthor{\bsnm{Wu}, \binits{Z.}},
\oauthor{\bsnm{Qi}, \binits{Y.}},
\oauthor{\bsnm{Li}, \binits{W.}},
\oauthor{\bsnm{Guan}, \binits{Y.}},
\oauthor{\bsnm{Zeng}, \binits{K.}},
\oauthor{\bsnm{Qi}, \binits{J.}},
\oauthor{\bsnm{Jin}, \binits{H.}},
\oauthor{\bsnm{Liu}, \binits{J.}},
\oauthor{\bsnm{Gu}, \binits{Y.}},
\oauthor{\bsnm{Yao}, \binits{Y.}},
\oauthor{\bsnm{Ding}, \binits{N.}},
\oauthor{\bsnm{Hou}, \binits{L.}},
\oauthor{\bsnm{Liu}, \binits{Z.}},
\oauthor{\bsnm{Xu}, \binits{B.}},
\oauthor{\bsnm{Tang}, \binits{J.}},
\oauthor{\bsnm{Li}, \binits{J.}}:
Kola: Carefully benchmarking world knowledge of large language models.
ICLR
(2024)
\end{botherref}
\endbibitem

%%% 5
\bibitem[\protect\citeauthoryear{Meo et~al.}{2023}]{medical_science}
\begin{botherref}
\oauthor{\bsnm{Meo}, \binits{S.A.}},
\oauthor{\bsnm{Al-Masri}, \binits{A.A.}},
\oauthor{\bsnm{Alotaibi}, \binits{M.}},
\oauthor{\bsnm{Meo}, \binits{M.Z.S.}},
\oauthor{\bsnm{Meo}, \binits{M.O.S.}}:
Chatgpt knowledge evaluation in basic and clinical medical sciences: Multiple choice question examination-based performance.
Healthcare (Basel).
(2023)
\end{botherref}
\endbibitem

%%% 6
\bibitem[\protect\citeauthoryear{Sennrich et~al.}{2016}]{bpe}
\begin{botherref}
\oauthor{\bsnm{Sennrich}, \binits{R.}},
\oauthor{\bsnm{Haddow}, \binits{B.}},
\oauthor{\bsnm{Birch}, \binits{A.}}:
Neural machine translation of rare words with subword units.
Proceedings of the 54th Annual Meeting of the Association for Computational Linguistics
(2016)
\end{botherref}
\endbibitem

%%% 7
\bibitem[\protect\citeauthoryear{Gilardi et~al.}{2023}]{zero-shot}
\begin{botherref}
\oauthor{\bsnm{Gilardi}, \binits{F.}},
\oauthor{\bsnm{Alizadeh}, \binits{M.}},
\oauthor{\bsnm{Kubli}, \binits{M.}}:
Chatgpt outperforms crowd-workers for text-annotation tasks.
Proceedings of the National Academy of Sciences
(2023)
\end{botherref}
\endbibitem

%%% 8
\bibitem[\protect\citeauthoryear{Sarkar et~al.}{2022}]{zero-shot-emnlp2022}
\begin{botherref}
\oauthor{\bsnm{Sarkar}, \binits{S.}},
\oauthor{\bsnm{Feng}, \binits{D.}},
\oauthor{\bsnm{Santu1}, \binits{S.K.K.}}:
Zero-shot multi-label topic inference with sentence encoders \& llms.
EMNLP
(2022)
\end{botherref}
\endbibitem

%%% 9
\bibitem[\protect\citeauthoryear{Osnabrügge et~al.}{2021}]{cross_domain}
\begin{botherref}
\oauthor{\bsnm{Osnabrügge}, \binits{M.}},
\oauthor{\bsnm{Ash}, \binits{E.}},
\oauthor{\bsnm{Morelli}, \binits{M.}}:
{Cross-Domain Topic Classification for Political Texts}.
Political Analysis
(2021)
\end{botherref}
\endbibitem

%%% 10
\bibitem[\protect\citeauthoryear{Bonica}{2014}]{bonica}
\begin{barticle}
\bauthor{\bsnm{Bonica}, \binits{A.}}:
\batitle{Mapping the ideological marketplace}.
\bjtitle{American Journal of Political Science}
\bvolume{58}(\bissue{2}),
\bfpage{367}--\blpage{386}
(\byear{2014})
\end{barticle}
\endbibitem

%%% 11
\bibitem[\protect\citeauthoryear{Box-Steffensmeier et~al.}{2013}]{amicus}
\begin{barticle}
\bauthor{\bsnm{Box-Steffensmeier}, \binits{J.M.}},
\bauthor{\bsnm{Christenson}, \binits{D.P.}},
\bauthor{\bsnm{Hitt}, \binits{M.P.}}:
\batitle{Quality over quantity: Amici influence and judicial decision making}.
\bjtitle{American Political Science Review}
\bvolume{107}(\bissue{3}),
\bfpage{446}--\blpage{460}
(\byear{2013})
\end{barticle}
\endbibitem

%%% 12
\bibitem[\protect\citeauthoryear{Muchlinski et~al.}{2016}]{predicted_acc1}
\begin{barticle}
\bauthor{\bsnm{Muchlinski}, \binits{D.}},
\bauthor{\bsnm{Siroky}, \binits{D.}},
\bauthor{\bsnm{He}, \binits{J.}},
\bauthor{\bsnm{Kocher}, \binits{M.}}:
\batitle{{Comparing Random Forest with Logistic Regression for Predicting Class-Imbalanced Civil War Onset Data}}.
\bjtitle{Political Analysis}
\bvolume{24}(\bissue{1}),
\bfpage{87}--\blpage{103}
(\byear{2016})
\end{barticle}
\endbibitem

%%% 13
\bibitem[\protect\citeauthoryear{Wang}{2019}]{predicted_acc2}
\begin{barticle}
\bauthor{\bsnm{Wang}, \binits{Y.}}:
\batitle{{Comparing Random Forest with Logistic Regression for Predicting Class-Imbalanced Civil War Onset Data: A Comment}}.
\bjtitle{Political Analysis}
\bvolume{21}(\bissue{1}),
\bfpage{107}--\blpage{110}
(\byear{2019})
\end{barticle}
\endbibitem

\end{thebibliography}

\end{document}